\documentclass[conference,9pt]{IEEEtran}
\IEEEoverridecommandlockouts
\usepackage{cite}
\usepackage{amsmath,amssymb,amsfonts}
\usepackage{algorithmic}
\usepackage{graphicx}
\usepackage{textcomp}
\usepackage{xcolor}
\def\BibTeX{{\rm B\kern-.05em{\sc i\kern-.025em b}\kern-.08em
    T\kern-.1667em\lower.7ex\hbox{E}\kern-.125emX}}

\usepackage{graphicx}
\usepackage[utf8]{inputenc} 
\usepackage[T1]{fontenc}    
\usepackage{hyperref}       
\usepackage{url}            
\usepackage{booktabs}       
\usepackage{amsfonts}       
\usepackage{nicefrac}       
\usepackage{microtype}      
\usepackage{xcolor}         
\usepackage{wrapfig}

\usepackage{booktabs} 

\usepackage{hyperref}

\usepackage{url}
\usepackage{graphicx}
\usepackage{amsmath}
\usepackage{amsthm}
\usepackage{booktabs}
\usepackage{algorithm}
\usepackage{algorithmic}
\usepackage{booktabs}
\usepackage{multirow}
\usepackage[utf8]{inputenc}
\usepackage{tcolorbox}

\usepackage[capitalize,noabbrev]{cleveref}

\theoremstyle{plain}
\newtheorem{theorem}{Theorem}[section]

\theoremstyle{definition}
\newtheorem{definition}[theorem]{Definition}

\theoremstyle{remark}

\begin{document}

\title{PracMHBench: Re-evaluating Model-Heterogeneous Federated Learning Based on Practical Edge Device Constraints}

\author{\IEEEauthorblockN{Yuanchun Guo, Bingyan Liu*\thanks{*corresponding author. }, Yulong Sha, Zhensheng Xian \thanks{This work was partly supported by the National Natural Science Foundation of China (62302054) and CCF-Huawei Populus Euphratica Fund (System Software Track, CCF-HuaweiSY202410).}}
\IEEEauthorblockA{\{gyc2001, bingyanliu, shayulong, xsin\}@bupt.edu.cn} 
\textit{School of Computer Science} ,
\textit{Beijing University of Posts and Telecommunications},
Beijing, China 
}
\maketitle

\begin{abstract}
Federating heterogeneous models on edge devices with diverse resource constraints has been a notable trend in recent years. 
Compared to traditional federated learning (FL) that assumes an identical model architecture to cooperate, model-heterogeneous FL is more practical and flexible since the model can be customized to satisfy the deployment requirement. Unfortunately, no prior work ever dives into the existing model-heterogeneous FL algorithms under the practical edge device constraints and provides quantitative analysis on various data scenarios and metrics, which motivates us to rethink and re-evaluate this paradigm. In our work, we construct the first system platform \textbf{PracMHBench} to evaluate model-heterogeneous FL on practical constraints of edge devices, where diverse model heterogeneity algorithms are classified and tested on multiple data tasks and metrics. Based on the platform, we perform extensive experiments on these algorithms under the different edge constraints to observe their applicability and the corresponding heterogeneity pattern.
\end{abstract}


\section{Introduction}


Federated learning (FL) is empowering multiple real-world 
applications such as Google’s Keyboard \cite{bonawitz2019towards} and medical image analysis \cite{liu2021feddg}, whose key idea is to employ a set of personal clients to collaboratively train a deep learning model with the orchestration of a central server. In recent years, conducting FL among edge devices has become a noteworthy trend \cite{liu2021distfl,li2022pyramidfl}. Under this condition, the typical FL assumption with respect to keeping an identical model architecture is hard to hold since edge devices are usually resource-constrained.

The research community has recognized this issue and introduced the concept of \textit{Model-Heterogeneous Federated Learning (MHFL)} \cite{diao2020heterofl}. In this context, each client maintains a model that accounts for the device's resource constraints, and the federated learning process operates among these heterogeneous models. MHFL can offer the following benefits: (1) For resource-constrained devices, MHFL provides the flexibility to adaptively conduct FL under diverse resource constraints (e.g., computation, communication, and memory) by adjusting the model size. (2) For individual edge users with distinctive preferences, MHFL allows them to customize the model's structure and size. Overall, MHFL is well-suited for deployment on various edge devices.

In the past few years, a large number of papers have been published to improve the performance of MHFL \cite{alam2022fedrolex,tan2022fedproto}. After exploration, we find that all of them are based on simple configurations (i.e., using the proportion of the original model to satisfy device constraints). However, they do not consider real-world device constraints. As shown in Table \ref{tab:diff}, on two real edge devices (Jetson Nano, Jetson Orin NX), models generated by different heterogeneous methods with the same proportion exhibit significant differences in terms of various resources, particularly for the computational cost (training time) and the memory usage. Therefore, \textit{we argue that the current comparisons of model-heterogeneous federated learning methods are not fair, and the community should rethink and re-evaluate model-heterogeneous federated learning under practical device constraints}.


To gain a fair and comprehensive performance comparison of diverse model-heterogeneous mechanisms for real-world constraints of edge devices, we construct the first system platform \textbf{PracMHBench}. PracMHBench defines three representative heterogeneity levels, each of which includes several corresponding MHFL algorithms. The platform also contains multiple model architectures and metrics, which are evaluated on various data tasks, spanning from computer vision (CV), natural language processing (NLP), and human activity recognition (HAR). Based on the benchmark, PracMHBench further build three typical edge-side constraint cases,
which we call \textit{computation-limited MHFL, communication-limited MHFL, and memory-limited MHFL}, and conducts experiments on them. For each limitation setting, we create it based on the collected statistics of practical edge devices, such that the results can benefit and guide the MHFL deployment in real-world resource-constrained scenarios.


Based on PracMHBench, we first perform extensive experiments on different model-heterogeneous algorithms under the specific device constraint to demystify the performance on different data tasks. Next, in terms of our evaluated metrics, we analyze the pattern of different heterogeneity algorithms and levels, aiming to explore the most suitable strategy for different resource-constrained scenarios. Finally, through the experiments, we summarize several insightful observations and conclusions (details in Section \ref{sec:experiment}). 

This paper makes the following contributions:
\begin{itemize}
    \item We construct a system platform PracMHBench to characterize and evaluate diverse model-heterogeneous algorithms under the practical resource-constrained edge devices. To the best of our knowledge, this is the first work that delves into the deployment of MHFL on real-world scenarios \footnote{https://github.com/9yc/PracMHBench}.
    
    \item We perform extensive experiments on PracMHBench with different heterogeneity levels and device constraints, and the results reveal a comprehensive landscape of current MHFL algorithms from a real system perspective. 
    
    \item Based on the experiments, we summarize the insightful observations and practical implications that can benefit the practitioners and researchers targeting resource-constrained MHFL systems.
\end{itemize}

\begin{table}[t]
\small
\caption{Models generated by different heterogeneous methods with the same proportion (i.e., x0.5 of the original model) on Jetson Orin NX and Jeston Nano. The statistics include the number of parameters, training time (one round) and memory usage. 'N' represents the Jetson Nano. 'O' represents the Jetson Orin NX.}
\label{tab:diff}
\centering
\begin{tabular}{ccccc}
\toprule
\multicolumn{1}{c}{Method} & \multicolumn{1}{c}{Model} & \multicolumn{1}{c}{Parameters(M)} & \multicolumn{1}{c}{\begin{tabular}[c]{@{}c@{}}Training\\ Time(s)\end{tabular}} & \multicolumn{1}{c}{\begin{tabular}[c]{@{}c@{}}Memory\\ Usage(M)\end{tabular}} \\
  \midrule
\multirow{2}{*}{SHeteroFL} & \multirow{2}{*}{\begin{tabular}[c]{@{}c@{}}ResNet101\\ (x0.5)\end{tabular}} & \multirow{2}{*}{10.66} & N:430.24 & \multirow{2}{*}{593}  \\
                           &                                                                        &                        & O:212.72 &                       \\ \midrule
\multirow{2}{*}{DepthFL}   & \multirow{2}{*}{\begin{tabular}[c]{@{}c@{}}ResNet101\\ (x0.5)\end{tabular}} & \multirow{2}{*}{10.29} & N:515.93 & \multirow{2}{*}{1220} \\
                           &                                                                        &                        & O:254.65 &                       \\ \midrule
\multirow{2}{*}{FedRolex}  & \multirow{2}{*}{\begin{tabular}[c]{@{}c@{}}ResNet101\\ (*0.5)\end{tabular}} & \multirow{2}{*}{10.75} & N:465.17 & \multirow{2}{*}{780}  \\
                           &                                                                        &                        & O:233.56 &                       \\ \midrule
\multirow{2}{*}{FeDepth}  & \multirow{2}{*}{\begin{tabular}[c]{@{}c@{}}ResNet101\\ (*0.5)\end{tabular}} & \multirow{2}{*}{10.54} & N:450.64 & \multirow{2}{*}{631}  \\
                           &                                                                        &                        & O:222.35 &                       \\\bottomrule
\end{tabular}
\vskip -0.2in
\end{table}

\begin{table*}
\small
\caption{Statistics of the proposed platform. Here `Hetero' refers to the heterogeneity level.}
\label{tab:statisitics}
\vskip -0.1in
\centering
\resizebox{0.9\linewidth}{!}{
\begin{tabular}{cc|cc|cc|cc}
\toprule
\multirow{2}{*}{\textbf{Hetero}} & \multirow{2}{*}{\textbf{Algorithm}} & \multicolumn{2}{c|}{\textbf{CV}}                                                                                                                                                                                              & \multicolumn{2}{c|}{\textbf{NLP}}                                                                                                                                                                                                              & \multicolumn{2}{c}{\textbf{HAR}}                                                                                                                                           \\ \cline{3-8} 
                                 &                                     & \textbf{Model}                                                                                                                                 & \textbf{Dataset}                                                             & \textbf{Model}                                                                                                                                         & \textbf{Dataset}                                                                      & \textbf{Model}                                                                               & \textbf{Dataset}                                                            \\ \midrule
\multirow{3}{*}{Width}           & Fjord                               & \multicolumn{1}{c|}{\multirow{6}{*}{\begin{tabular}[c]{@{}c@{}}ResNet-101\\ MobileNetV2 \\ Variants\\ (100\%,75\%,\\ 50\%,25\%)\end{tabular}}} & \multirow{8}{*}{\begin{tabular}[c]{@{}c@{}}CIFAR10,\\ CIFAR100\end{tabular}} & \multicolumn{1}{c|}{\multirow{6}{*}{\begin{tabular}[c]{@{}c@{}}ALBERT,\\ Customized \\ Transformer\\ Variants\\ (100\%,75\%,\\ 50\%,25\%)\end{tabular}}} & \multirow{8}{*}{\begin{tabular}[c]{@{}c@{}}Stack-\\ Overflow,\\ AG-news\end{tabular}} & \multicolumn{1}{c|}{\multirow{8}{*}{\begin{tabular}[c]{@{}c@{}}Customized\\ CNN\end{tabular}}} & \multirow{8}{*}{\begin{tabular}[c]{@{}c@{}}HAR-BOX,\\ UCI-HAR\end{tabular}} \\ \cline{2-2}
                                 & SHeteroFL                           & \multicolumn{1}{c|}{}                                                                                                                          &                                                                              & \multicolumn{1}{c|}{}                                                                                                                                  &                                                                                       & \multicolumn{1}{c|}{}                                                                        &                                                                             \\ \cline{2-2}
                                 & FedRolex                            & \multicolumn{1}{c|}{}                                                                                                                          &                                                                              & \multicolumn{1}{c|}{}                                                                                                                                  &                                                                                       & \multicolumn{1}{c|}{}                                                                        &                                                                             \\ \cline{1-2}
\multirow{3}{*}{Depth}           & FeDepth                             & \multicolumn{1}{c|}{}                                                                                                                          &                                                                              & \multicolumn{1}{c|}{}                                                                                                                                  &                                                                                       & \multicolumn{1}{c|}{}                                                                        &                                                                             \\ \cline{2-2}
                                 & InclusiveFL                         & \multicolumn{1}{c|}{}                                                                                                                          &                                                                              & \multicolumn{1}{c|}{}                                                                                                                                  &                                                                                       & \multicolumn{1}{c|}{}                                                                        &                                                                             \\ \cline{2-2}
                                 & DepthFL                             & \multicolumn{1}{c|}{}                                                                                                                          &                                                                              & \multicolumn{1}{c|}{}                                                                                                                                  &                                                                                       & \multicolumn{1}{c|}{}                                                                        &                                                                             \\ \cline{1-2} \cline{5-5}
\multirow{2}{*}{Topology}        & FedProto                            & \multicolumn{1}{c|}{\multirow{2}{*}{\begin{tabular}[c]{@{}c@{}}MobileNet,\\ Resnet Family\end{tabular}}}                                       &                                                                              & \multicolumn{1}{c|}{\multirow{2}{*}{Albert Family}}                                                                                                    &                                                                                       & \multicolumn{1}{c|}{}                                                                        &                                                                             \\ \cline{2-2}
                                 & Fed-ET                              & \multicolumn{1}{c|}{}                                                                                                                          &                                                                              & \multicolumn{1}{c|}{}                                                                                                                                  &                                                                                       & \multicolumn{1}{c|}{}                                                                        &                                                                             \\ \bottomrule
\end{tabular}}
\end{table*}

\section{Background and Related Work}



\textbf{Model Heterogeneity in Federated Learning.}
In recent years, Federated learning (FL) tailored for massive edge devices has garnered unprecedented attention \cite{liu2021distfl,shin2022fedbalancer,li2022pyramidfl, cai2023federated, liu2023beyond}. Unlike traditional federated learning that only focuses on algorithm performance, federated learning in such scenarios pays more attention to the characteristics of heterogeneous devices. For example, training VGG16 requires over 8GB memory with a batch size of 8 \cite{deng2022tailorfl}, making it hard to deploy for many edge devices. The research community has proposed \textit{model-heterogeneous federated learning (MHFL)} to address the issue. The primary aim is to enable each device to customize its model size while ensuring federation with other devices.



Achieving MHFL encompasses two categories of state-of-the-art techniques. The first category is sub-model partial aggregation \cite{diao2020heterofl, alam2022fedrolex, horvath2021fjord}. Its fundamental idea involves extracting sub-models from a large model that adheres to the memory constraints of devices, followed by weight aggregation based on the positional information of neurons. 
The second category employs knowledge distillation to achieve federated aggregation \cite{he2020group, cho2022heterogeneous,zhang2022fedduap, tan2022fedproto}. Its primary pipeline involves: (1) The server leverages a public dataset to compute the output logits of each model and performs logits aggregation; (2) The server sends the aggregated logits back to each client, thereby enabling them as the teacher model for the distillation process. 
While MHFL has been extensively studied, its level of heterogeneity and the performance of various methods under different real device constraints  have not been thoroughly explored. Our work aims to fill this gap by constructing a comprehensive and practical platform for MHFL in resource-constrained edge systems.

\textbf{FL Platforms.}
Benchmarking and evaluating the performance of FL is essential for practitioners to understand and employ associated algorithms and configurations effectively. At present, there are two main categories of platforms for FL. The first category comprises general FL platforms, which primarily aim to comprehensively assess the performance of various FL algorithms under different conditions \cite{caldas2018leaf, chai2020fedeval, gao2020end}.  
The second category consists of specific FL platforms, tailored for in-depth exploration of certain characteristics within FL systems \cite{lai2022fedscale, li2022federated, yang2021characterizing}. For example, to assess the data heterogeneity, Li \textit{et al.} \cite{li2022federated} proposed a comprehensive data partitioning strategy to cover typical non-IID (non-independently and identically distributed) data cases and evaluated existing FL algorithms under such settings. Furthermore, Yang \textit{et al.} \cite{yang2021characterizing} developed an FL platform to explore the impact of heterogeneity on performance, particularly focusing on hardware and state heterogeneity. 
In our study, we pay more attention to the impact of \textit{model heterogeneity} on FL performance and the practice of conducting diverse heterogeneous methods on resource-constrained edge devices, which is fundamentally different from previous research efforts.


\section{Benchmark Construction}

\textbf{PracMHBench} is a platform aimed to characterize and evaluate diverse model heterogeneity algorithms on practical resource-constrained devices. In this section, we first describe its benchmark construction in detail. The concrete statistics are shown in Table \ref{tab:statisitics}.



\begin{figure}[htbp]
\centering
\includegraphics[width=1\columnwidth]{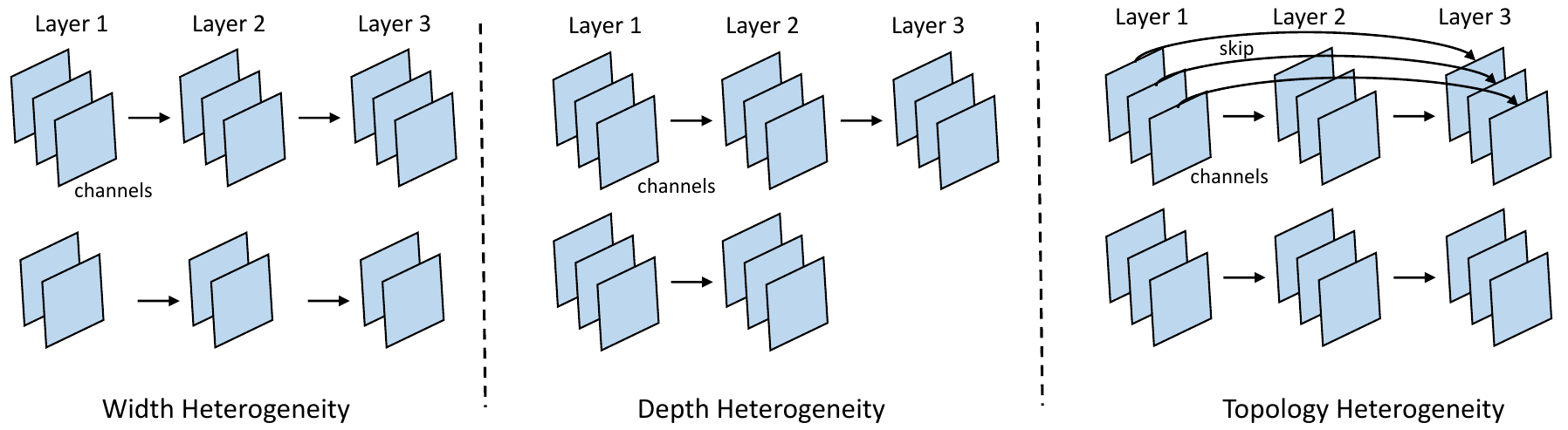} 
\vskip -0.1in
\caption{Demonstration of different heterogeneity levels. Here `skip' denotes the skip connection used in ResNets.}
\label{fig:level}

\end{figure}

\textbf{Model Heterogeneity Levels.} Due to diverse resource constraint conditions, various clients in federated learning may equip models with different capacities or architectures. In our PracMHBench, we define and categorize the model heterogeneity mechanism in an FL system into three levels (as shown in Figure \ref{fig:level}):
\begin{itemize}
    \item \textit{Width Heterogeneity}, where the models employed by clients share the same topology while exhibiting differences in the number of channels in each layer, resulting in heterogeneous models with various widths.
    
    \item \textit{Depth Heterogeneity}, where the model topology employed by clients is also identical and the difference lies in the number of layers, which can be considered as the depth-level heterogeneity.
    

    \item \textit{Topology Heterogeneity}, where clients employ models with entirely different topologies. For example, an FL system may consist of diverse model architectures, such as ResNet \cite{he2016deep}, EfficientNet \cite{tan2019efficientnet}, MobileNet \cite{howard2017mobilenets}, and GoogleLeNet \cite{szegedy2015going}, in order to satisfy various user preferences and requirements.
\end{itemize}

\textbf{Representative Heterogeneous Algorithms.} 
In terms of the aforementioned heterogeneity levels, we select several heterogeneous algorithms for each level to represent its performance. Specifically, for \textit{width heterogeneity}, we adopt Fjord \cite{horvath2021fjord}, SHeteroFL \cite{diao2020heterofl} and FedRolex \cite{alam2022fedrolex}, whose key settings are allocating sub-models with different widths to clients before conducting specifically designed federation schemes. In the case of \textit{depth heterogeneity}, we select  FeDepth \cite{zhang2023memory}, InclusiveFL \cite{liu2022no} and DepthFL \cite{kim2022depthfl}, which achieves depth scaling by pruning one or more of the top layers of the global model. Towards \textit{topology heterogeneity}, FedProto \cite{tan2022fedproto} and Fed-ET \cite{cho2022heterogeneous} are used to conduct FL with the help of knowledge distillation among these heterogeneous architectures.

\textbf{Data Tasks.}
In our platform, we target three areas: computer vision (CV), natural language processing (NLP), and human activity recognition (HAR), each of which is evaluated by two representative datasets. For CV, we utilize CIFAR-10 and  CIFAR100 \cite{krizhevsky2009learning}. For NLP, we employ AG-News \cite{zhang2015character} and Stack-OverFlow \cite{tff}, and for HAR, we choose HAR-BOX \cite{ouyang2021clusterfl} and UCI-HAR \cite{anguita2013public}.

\textbf{Model Architectures.}
To evaluate the different levels of model heterogeneity, we select various model architectures for experiments. For width and depth heterogeneity, we follow the settings outlined in related works \cite{diao2020heterofl,alam2022fedrolex}, conducting experiments using four different widths/depths (100\%, 75\%, 50\%, 25\% ) of the ResNet-101  \cite{he2016deep} on the CIFAR-100 dataset \cite{krizhevsky2009learning} and MobileNetV2 \cite{howard2017mobilenets} on CIFAR-10 dataset\cite{krizhevsky2009learning}. Additionally, for NLP tasks, we employ a customized transformer model \cite{vaswani2017attention} on AG-news  and ALBERT \cite{lan2019albert} on Stack Overflow. We partition them into various dimensions to meet different definitions of heterogeneity. For HAR tasks, we employ customized CNNs following the previous work \cite{ek2021evaluating} and partition it into various dimensions to meet different definitions of heterogeneity.
Regarding topology heterogeneity, we conduct experiments using two model families with distinctive architectures on CV tasks, including the ResNet Family \cite{he2016deep} (ResNet-18, ResNet-34, ResNet-50, ResNet-101) on CIFAR-100 \cite{krizhevsky2009learning},  and the MobileNet Family \cite{howard2017mobilenets} (MobileNetV2, MobileNetV3 small, MobileNetV3 large) on CIFAR-10 \cite{krizhevsky2009learning}. Considering  NLP models, we choose the ALBERT Family \cite{lan2019albert} (ALBERT base, ALBERT large, ALBERT xxlarge) on Stack Overflow \cite{tff}. Additionally, we modified the structure of the customized CNNs  for HAR tasks.



\textbf{Evaluated Metrics.}
To provide a comprehensive evaluation, PracMHBench adopts the following four metrics: 
(i) \textit{Global accuracy}, where we maintain a global dataset and the final federated model is evaluated on it. 
(ii) \textit{Time-to-accuracy}, which is defined as the wall clock time for training a deep learning model to reach a pre-set accuracy, aiming at measuring the training speed.
(iii) \textit{Stability}, where we record the accuracy of each device and compute the variance among them to show the stability for various heterogeneous models.
(iv) \textit{Effectiveness}, where we compare the final accuracy of MHFL algorithms with a simple resource-aware homogeneous baseline (i.e., training the smallest homogeneous model across all heterogeneous devices). The effectiveness is measured by the accuracy improvement. Our goal is to test whether heterogeneous models are actually needed (i.e., whether the improvement is greater than zero) and the benefit of conducting MHFL (i.e., the improvement value).



\begin{figure*}[t]
\centering
\includegraphics[width=2\columnwidth]{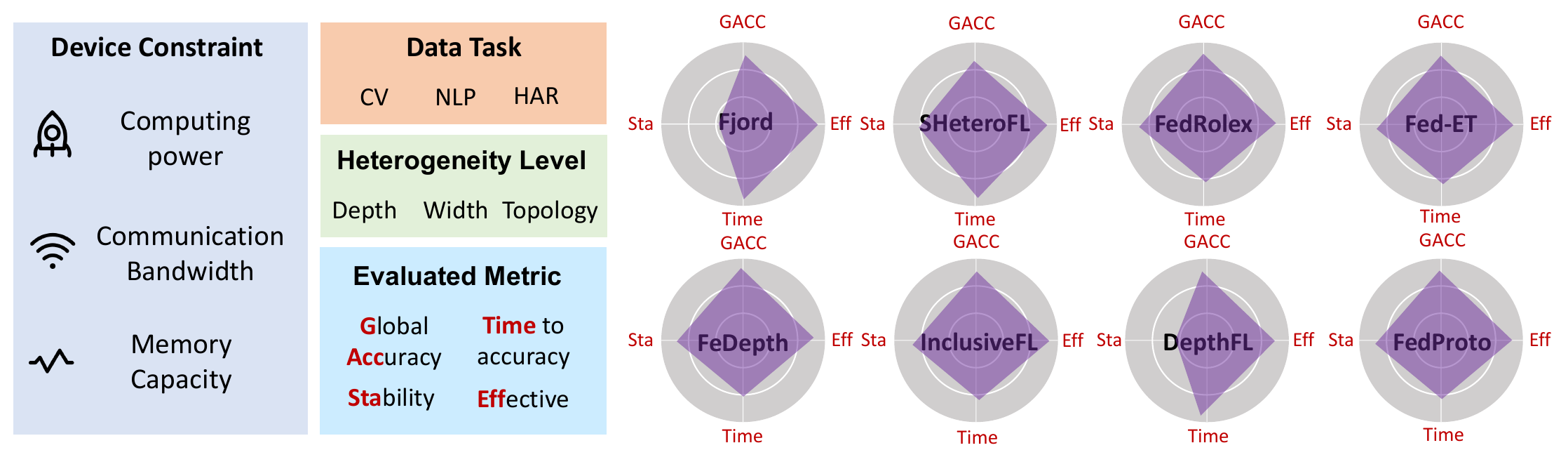} 
\vskip -0.1in
\caption{Evaluation track of our platform PracMHBench. Here the information in radar charts is just for demonstration. The concrete performance will be compared in the subsequent experiments.}
\label{fig:overview}
\end{figure*}

\begin{figure}[htbp]
\centering
\includegraphics[width=1\columnwidth]{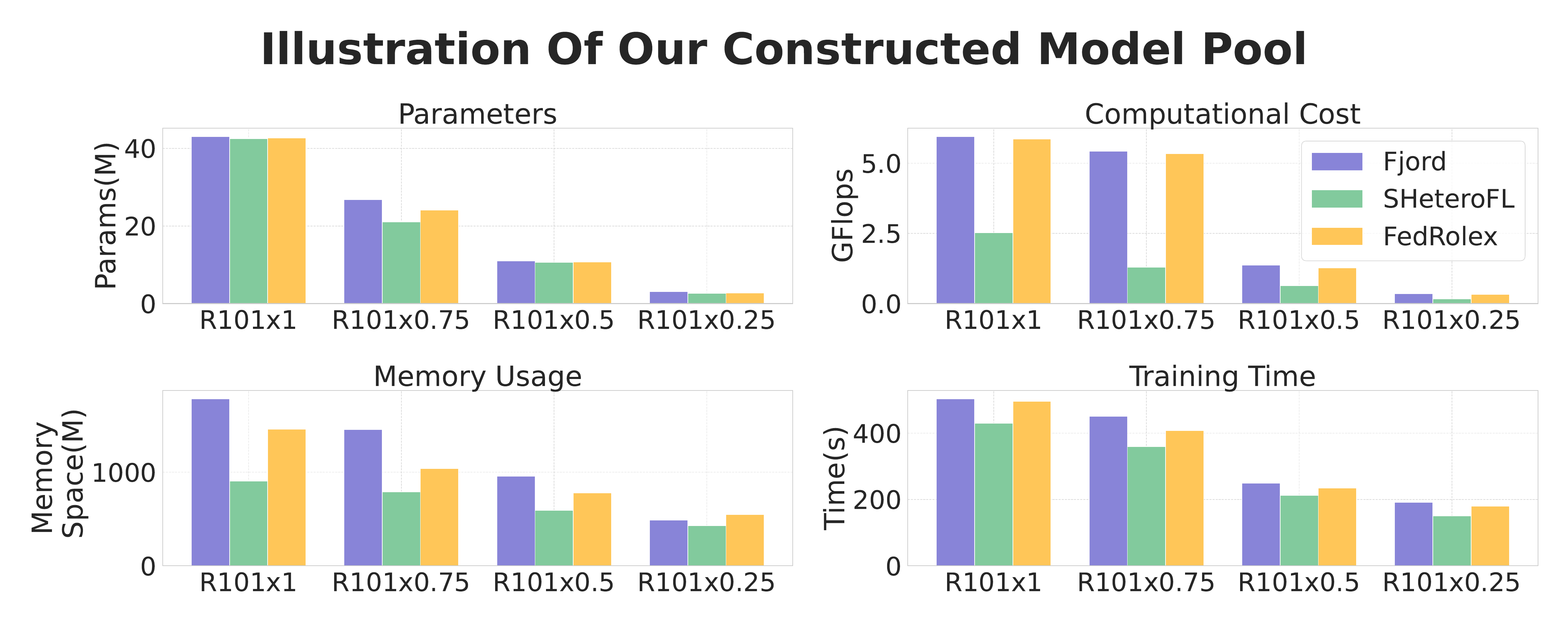} 
\vskip -0.2in
\caption{Illustration of our constructed model pool. We use the tested system statistics (e.g., Parameters, GFlops) for three algorithms on Jetson Orin NX as an example. 'R' refers to ResNet. }
\label{fig:pool}
\vskip -0.1in
\end{figure}

\begin{table}[t]
\centering
\caption{Edge devices used in our platform construction.}
\vskip -0.1in
\label{tab:boards}
\begin{tabular}{c|c|c}
\toprule
\textbf{Devices} & \textbf{Processors}           &\textbf{GPU Memory}                                                    \\ \midrule
Jetson orin nx   & 1024-core NVIDIA Ampere GPU       &16GB                                                              \\ \midrule
Jetson tx2 nx    & 256-core NVIDIA Pascal GPU        &4GB                                                              \\ \midrule
Raspberry 4B     & \begin{tabular}[c]{@{}c@{}}Broadcom BCM2711B0 quad-core \\ A72 64-bit @ 1.5GHz CPU\end{tabular}  & no gpu\\ \bottomrule
\end{tabular}
\vskip -0.1in

\end{table}

\section{Practical Device Constraints}
Given the above benchmark settings, we then manually build three typical resource-constrained cases based on real-world device properties (enumerated in Table \ref{tab:boards}). As illustrated in Figure \ref{fig:pool},
we first build a model pool, where various deep learning models in the literature are measured and we pick out the suitable one in terms of the heterogeneity level and resource constraint. Our principle for selecting models from the model pool is to keep the constraint consistent for all methods to ensure fairness. In the following subsections, we describe each resource limitation on MHFL in detail. Note that for each specific limitation, other resources remain identical. 


\subsection{Case I: Computation-Limited MHFL}

\textbf{Description.} 
Limited computing power means that different devices may demonstrate varying levels of computational constraints. As shown in Table \ref{tab:diff}, there exists a significant difference in computing power among real edge devices, leading to distinctive training speed. Under this condition, we define \textit{Computation-Limited MHFL} as follows:

\begin{definition}
    \textbf{(Computation-Limited MHFL)}. 
    \textit{Given the various computing capabilities of practical edge devices, the goal of computation-limited MHFL is to adapt the model size of its original version to ensure a same model training speed among devices for synchronous aggregation.}
\end{definition}

\textbf{Setting.}
To achieve such a scenario, we follow the IMA dataset settings \cite{yang2021characterizing}, where they collected status from more than 1,000 devices (e.g., Samsung Note 10, Nexus 6, and Redmi Note 8) and recorded the real computational capabilities of these devices. We used the computing power of the IMA dataset to build the computation-limited scenarios. 
Specifically, we assign the diverse computational capabilities in the IMA dataset to the clients of the federated learning system as their training constraint, and select corresponding suitable models of different heterogeneity levels for each client from the model pool, with the help of the statistics of the training time for various models, which is provided by an AI Benchmark on various devices \cite{aibench}. This helps ensure roughly equivalent training times for each device and achieve synchronous aggregation under a fixed submission deadline.

\subsection{Case II: Communication-Limited MHFL}

\textbf{Description.}
The limited communication bandwidth indicates that the speed of uploading/downloading model parameters is affected by the actual device hardware and network environment. In an FL system, a client with small communication bandwidth may spend a lot of time for uploading or downloading model parameters, resulting in its inability to upload model parameters in time for aggregation. The \textit{Communication-Limited MHFL} can be defined as follows:

\begin{definition}
    \textbf{(Communication-Limited MHFL)}. 
    \textit{Given the various communication bandwidth of different devices, the goal of communication-limited MHFL is to adjust the model size of its original version to ensure a same uploading/downloading speed among devices for synchronous aggregation.}
\end{definition}


\textbf{Setting.}
In the case of communication constraints, we control the communication time of each round in the federated learning system to a certain time (e.g., 200s) based on the real network bandwidth information provided by the IMA dataset. We then select appropriate models and corresponding quantities from the model pool to deploy different heterogeneity levels. Note that the parameter sizes corresponding to the various models under the different methods vary, and thus the model proportion of different sizes in the different methods also varies.


\subsection{Case III: Memory-Limited MHFL}

\textbf{Description.}
Unlike the above, limited memory capacity focuses more on the actual memory requirements of training each model on different devices. Limited memory may have a direct impact on whether a client can conduct training and changing the model size can directly satisfy a specific memory limit. In this case, we define \textit{Memory-Limited MHFL} as follows:


\begin{definition}
    \textbf{(Memory-Limited MHFL)}. 
    \textit{Given the various memory capacity of different devices, the goal of memory-limited MHFL is to change the model size of its original version to ensure available local training for completing FL.}
\end{definition}

\textbf{Setting.} 
We choose three typical GPU memory sizes of real devices: 16GB (e.g., Jetson Orin nx), 4GB (e.g., Jetson tx2 nx), and no GPU (e.g., Raspberry Pi). We then conduct experiments on real devices to observe whether the models in the model pool can be trained for federated learning. The largest trainable model is assigned to the client for meeting the corresponding memory size. In addition, the client proportion of different memory sizes is determined by the real-world proportion distribution depicted in \cite{ram}.

\begin{figure*}
\centering
\includegraphics[width=1.8\columnwidth]{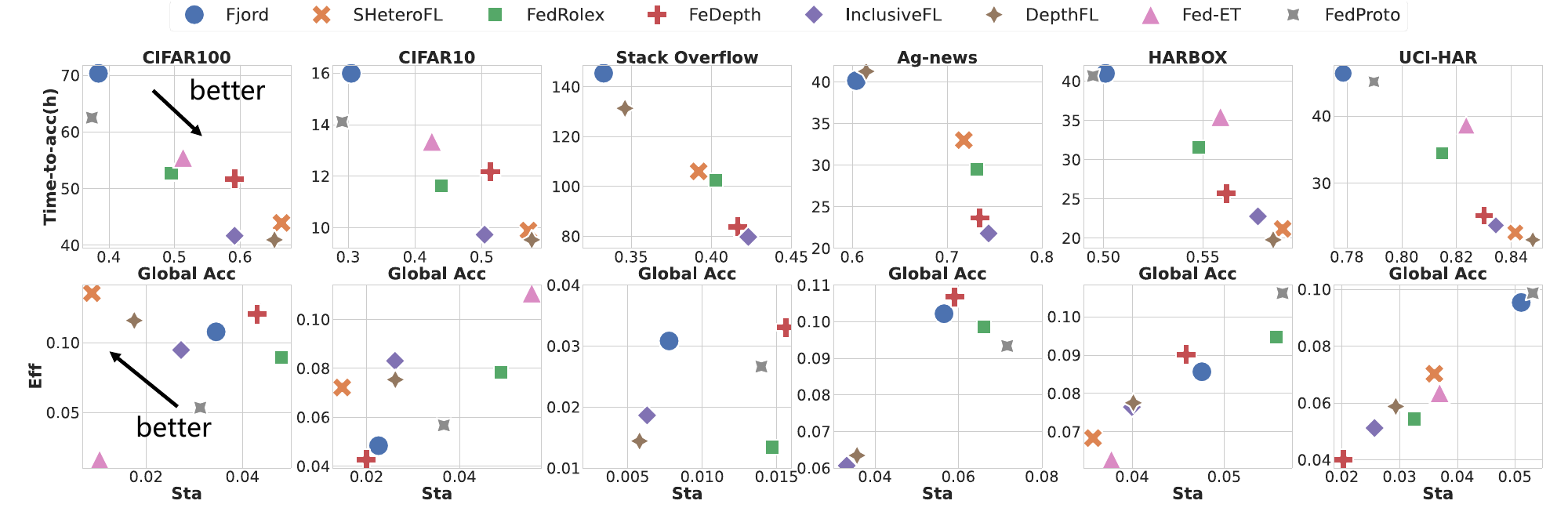} 
\vskip -0.2in
\caption{Results on computation-limited MHFL. The top row refers to the comprehensive performance of global accuracy and time-to-accuracy. The bottom row represents the performance of stability and effectiveness.}
\label{fig:compacc}
\vskip -0.in
\end{figure*}

\begin{figure*}
\centering
\includegraphics[width=1.8\columnwidth]{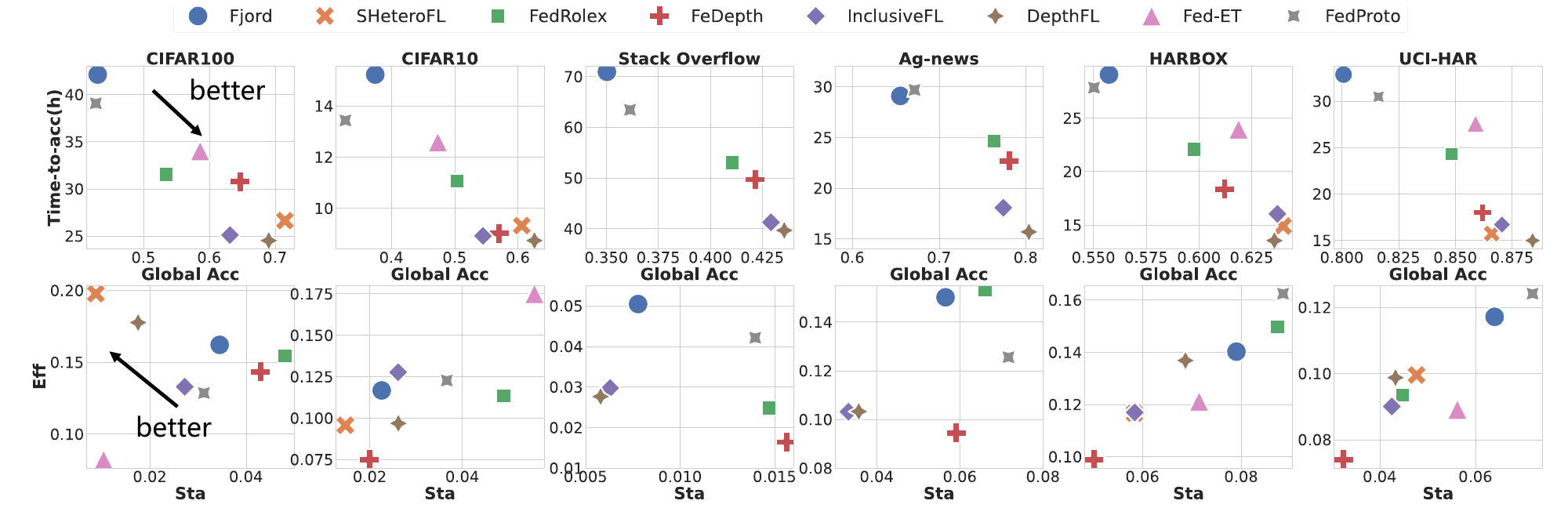} 
\vskip -0.2in
\caption{Results on communication-limited MHFL.}
\vskip -0.2in
\label{fig:commacc}
\end{figure*}

\begin{figure}[htbp]
\centering
\includegraphics[width=0.9\columnwidth]{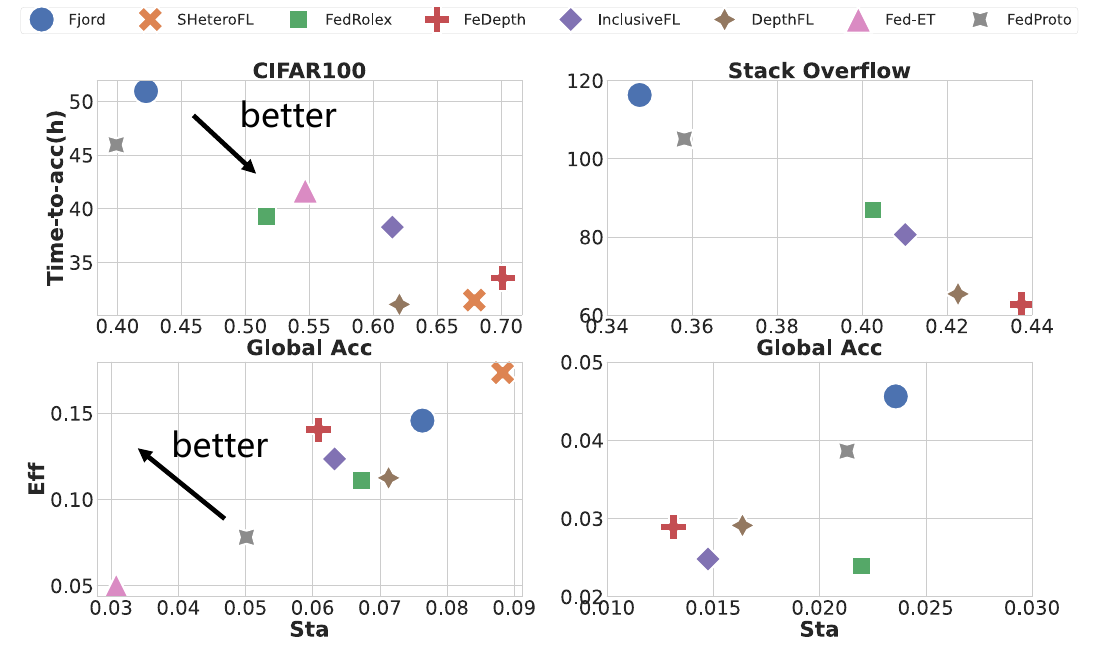} 
\vskip -0.2in
\caption{Results on memory-limited MHFL.}
\vskip -0.2in
\label{fig:memacc}
\end{figure}
\section{Evaluation and Analysis}
\label{sec:experiment}

Based on PracMHBench, we perform extensive experiments and analyze the results. The theme of this measurement is to quantitatively understand the performance discrepancy of different heterogeneity algorithms. As shown in Figure \ref{fig:overview}, the evaluation track includes: (1) selecting a targeted resource constraint; (2) conducting experiments on a variety of data tasks for different heterogeneity levels and algorithms; (3) recording and comparing the performance based on the evaluated metrics.
In the following parts, we will describe the results of each resource constraint in detail. Note that some methods are not compatible with NLP tasks, and thus we omit the corresponding results. In addition, for the memory-limited scenario, we only focus on large CNN models (i.e., Resnet-101) and Transformer models (i.e., ALBERT) since most of the current edge devices can hold small models for HAR tasks.

Specifically, for CIFAR10, CIFAR100 and AG-News, we adopt IID (Independent and Identically Distributed) partition. For Stack Overflow, HAR-BOX and UCI-HAR, the dataset is partitioned over user IDs, making the dataset naturally non-IID distributed. The number of clients for CIFAR10, CIFAR100, AG-News, Stack-OverFlow, HAR-BOX, and UCI-HAR is 100, 100, 50, 500, 100, 30. The sampling ratio is 10\%. All of the experiments are conducted for 1000 federated rounds to ensure convergence. Finally, we run each experiment 3 times and average them as the reported results.


\subsection{Results on Computation-Limited MHFL}
\label{sec:computation}

\textbf{Global accuracy and time-to-accuracy.}
The top row of Figure \ref{fig:compacc} shows the comprehensive results of global accuracy and time-to-accuracy. From the figure, we can observe that: (1) Overall, SHeteroFL and DepthFL achieve better performance on both global accuracy and time-to-accuracy. Regardless of the type of the data tasks, these two methods ensure relative superiority while Fjord and FedProto consistently exhibit the poorest performance across all data tasks. This suggests that under the computational constraint, the method's performance is unrelated to the type of data. (2) Although the two top-performing methods belong to different heterogeneity levels, we find that all depth-level methods (i.e., FeDepth, InclusiveFL, DepthFL) demonstrate favorable outcomes. This implies that adjusting the model size based on depth is a suitable choice in this scenario.

\textbf{Stability and effectiveness.}
The bottom row of Figure \ref{fig:compacc} demonstrates the performance of stability and effectiveness. The following observations can be drawn: (1) Different from global accuracy and time-to-accuracy, which can identify the best algorithm for all data tasks, achieving a simultaneous satisfaction of stability and effectiveness is challenging on most datasets except for CIFAR-10, where FeDepth can be considered the best performer, which means that we can enhance the performance across nearly all devices. For other datasets, these two metrics tend to exhibit a trade-off, meaning that as stability increases, effectiveness tends to decrease. (2) For different heterogeneous levels, their performance on these two metrics does not follow any specific pattern and varies according to the differences in datasets.

\begin{tcolorbox}[sharp corners]
\textbf{Observation}: \textit{In computation-limited MHFL, the superiority in terms of global accuracy and training speed remains a consistent pattern regardless of changes in data tasks. Among them, the methods belonging to the depth-level heterogeneity are relatively better than others. However, there is no one-size-fits-all algorithm that can achieve a desirable trade-off between the stability and effectiveness on diverse data tasks.}
\end{tcolorbox}



\subsection{Results on Communication-Limited MHFL}
\label{sec:communication}

\textbf{Global accuracy and time-to-accuracy.}
The top row of Figure \ref{fig:commacc} shows the comprehensive comparison. We can draw the following conclusions:
(1) Similar to the scenario of computational constraints, under communication bandwidth constraints, SHeteroFL and DepthFL still perform optimally, and this conclusion remains consistent in all data tasks. Besides, depth-level methods outperform other baselines. This indicates that the extent of changes in model size due to computational and communication constraints is not significant. Thus, the overall superiority of various model-heterogeneous algorithms remains unchanged.
(2) Although the performance ranking has not changed, there is still a noticeable variation in the training speed gap among different methods. For example, on the Stack Overflow dataset, under the communication bandwidth constraint, the gap of time-to-accuracy (i.e., training speed) is significantly smaller than it is in the computational constraint (roughly 2X). On other datasets, the conclusion also holds. We believe this is because the communication constraint is more related to the size of the model parameters, which better aligns with the original design principle of the algorithms (i.e., controlling the proportion of model size). Therefore, each method can essentially follow the original settings, and the fluctuation in training speed will not be too significant compared to the computational constraint, which not only considers the model parameters but also takes into account the computational capabilities of different devices.


\textbf{Stability and effectiveness.}
As depicted in the bottom row of Figure \ref{fig:commacc}, we record the stability and effectiveness performance under the communication limitation. We can see that: (1) At a high level, it is hard to figure out which algorithm can simultaneously ensure stability and effectiveness across all data tasks. In other words, these two metrics do not follow any specific pattern. (2) We discovered that heterogeneous methods at the depth level perform poorly in terms of effectiveness. The reason may be twofold: on the one hand, architectures related to depth itself exhibit good performance (with high global accuracy as mentioned in the figure); on the other hand, as stated above, the communication constraint does not differ significantly from proportional splitting. Therefore, the overall state aligns well with the original design of the corresponding paper. Therefore, the improvement after heterogeneity is not as pronounced.

\begin{tcolorbox}[sharp corners]
\textbf{Observation}: \textit{Communication-Limited MHFL somewhat resembles the proportional splitting in existing literature, as it pays more attention to the number of model parameters. Therefore, the gaps among various algorithms, especially in terms of training speed, are not significant. Additionally, compared to other heterogeneous levels, although depth-level heterogeneity still holds an advantage in global accuracy and training speed, the performance improvement (i.e., effectiveness) brought by heterogeneity is relatively limited.}
\end{tcolorbox}

\begin{figure}[htbp]
\centering
\includegraphics[width=1\columnwidth]{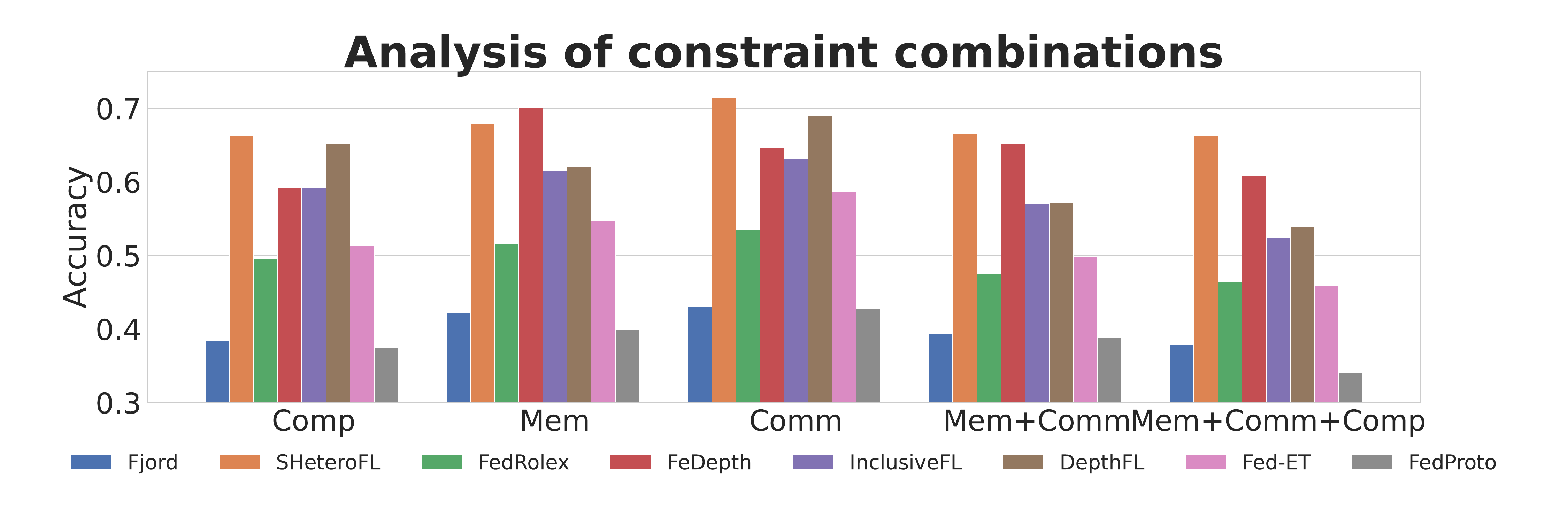} 
\vskip -0.2in
\caption{Analysis of constraint combinations.}
\label{fig:all}
\vskip -0.2in
\end{figure}

\begin{figure}[htbp]
\centering
\includegraphics[width=1\columnwidth]{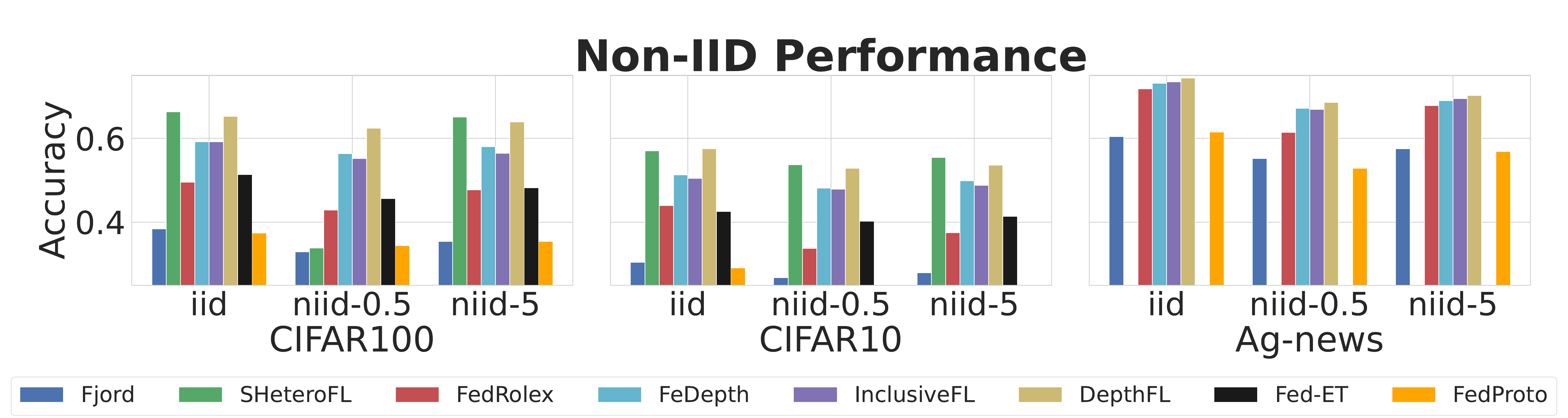} 
\vskip -0.1in
\caption{Non-iid performance on the computation-limited scenario. Here `0.5' and `5' are the alpha of the non-iid dirichlet partition.}
\label{fig:noniid}
\vskip -0.2in
\end{figure}

\begin{figure}[htbp]
\centering
\includegraphics[width=1\columnwidth]{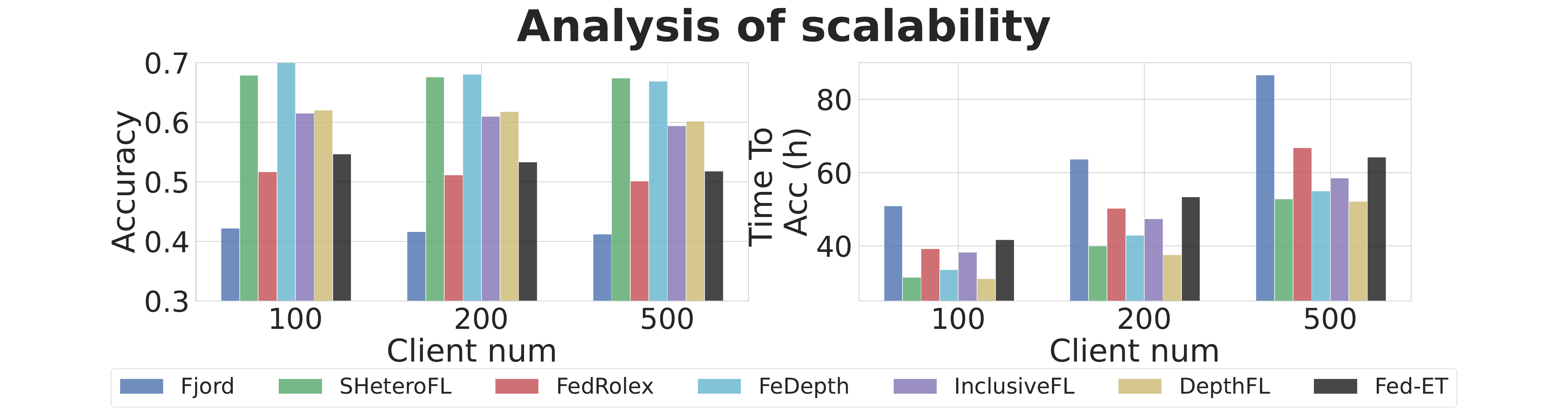} 
\vskip -0.1in
\caption{Analysis of scalability.}
\label{fig:scal}
\vskip -0.2in
\end{figure}

\subsection{Results on Memory-Limited MHFL}
\label{sec:memory}

\textbf{Global accuracy and time-to-accuracy.}
Figure \ref{fig:memacc} shows the results. From the top row, the following phenomenon is noticeable: We \textit{surprisingly} find that DepthFL, which performs well in the above two constraints, no longer exhibits superiority. Its global accuracy has significantly decreased, falling below the previously comparable SHeteroFL and inferior FeDepth. We attribute this to the fact that DepthFL itself requires a substantial amount of memory (shown in Table \ref{tab:diff}). When memory is constrained, DepthFL needs to drastically reduce its size, leading to a decrease in performance. Instead, although FeDepth does not perform well under the computational and communication constraints, its small memory footprint (shown in Table \ref{tab:diff}) allows it to accommodate larger models in this scenario, resulting in better performance.

\textbf{Stability and effectiveness.}
We show the results at the bottom of Figure \ref{fig:memacc}. From the figure, we find the following interesting observation: Compared to the two scenarios mentioned above, there is not much variation in the effectiveness trends of different algorithms. However, the conclusion regarding stability is almost reversed. For instance, in the case of CIFAR100, SHeteroFL exhibits the highest stability under the computation and communication constraints. However, for the memory constraint, its stability becomes the lowest among all methods. For Stack Overflow, Fjord demonstrates relatively high stability under the first two scenarios, but faced with memory limitations, its stability jumps to the lowest. We attribute this phenomenon to the significant difference between model parameters and actual device memory usage (as shown in Table \ref{tab:diff}). Therefore, under the memory constraint, the model size of many methods may undergo substantial changes, leading to unstable performance.

\begin{tcolorbox}[sharp corners]
\textbf{Observation}: \textit{In practical memory-constrained scenarios, the performance of global accuracy and training speed, apart from the influence of heterogeneous levels and the methods themselves, is significantly affected by the actual memory footprint occupied by the models of different methods. Moreover, it does not exhibit a strong correlation with the types of data tasks. Additionally, concerning the stability, although it remains challenging to identify specific patterns, its trends are precisely opposite to the conclusions drawn under the computation and communication constraints.}
\end{tcolorbox}

\subsection{Other Analysis}
\label{app:combinations}
\textbf{Analysis of constraint combinations.} Here we test two types of combinations: communication+memory limited FL and computation+communication+memory limited FL on the CIFAR-100 dataset. Figure \ref{fig:all} demonstrates the accuracy results.We can observe that: Under the combination of communication+memory, the performance of all methods decreases compared with the single restriction and SHeteroFL performs best when various constraints are combined together due to its excellent partitioning method. SHeteroFL's slimmable partition method is well controlled in terms of calculation volume, memory consumption and communication volume, making it more adaptable to various restricted environments. 

\textbf{Analysis of Non-IID Performance.} Figure \ref{fig:noniid} show the results of computational constraints under the non-iid scenario. From the figure, we can observe that although non-iid will degrade the performance, the conclusion is not changed, which
means that existing algorithms is robust to the non-iid scenario.

\textbf{Analysis of scalability.}
To demonstrate the scalability, here we conduct an experiment to evaluate the convergence-related metrics of various methods with different client numbers on the memory-limited MHFL setting and the CIFAR-100 dataset. As shown in Figure \ref{fig:scal}, scaling up leads to a decline in convergence performance, including both convergence effectiveness and speed. Among these methods, the width-level heterogeneity demonstrates relatively smaller performance losses, indicating better scalability.

\section{Conclusion}

In this work, we construct the first comprehensive platform for MHFL on practical resource constraints and conduct extensive measurements to quantitatively evaluate their performance. The results help reveal a complete landscape among diverse model heterogeneity levels and algorithms on real-edge devices. Based on the results, we summarize strong observations and implications that can be useful to FL developers and researchers of resource-constrained federated systems. 



 \medskip
{
\small
\bibliography{main}
\bibliographystyle{unsrt}
}






\clearpage

\end{document}